\documentclass[journal]{IEEEtran}
\ifCLASSINFOpdf
\else
\fi
\usepackage{booktabs}
\usepackage{amsmath}
\usepackage{amssymb}
\usepackage{graphicx}
\usepackage{adjustbox}
\usepackage{arydshln}
\usepackage{enumitem}
\usepackage{hyperref}
\usepackage{xcolor}
\usepackage{newtxtext,newtxmath}
\usepackage[numbers]{natbib}
\usepackage{multirow}

\usepackage[ruled,vlined,algo2e]{algorithm2e}
\usepackage{algorithm}
\usepackage{algorithmic}
\usepackage{subcaption}
\usepackage{graphicx}

\usepackage{tikz}
\usepackage[edges]{forest}
\definecolor{hiddendraw}{RGB}{205, 44, 36}
\definecolor{hidden-blue}{RGB}{194,232,247}
\definecolor{hidden-orange}{RGB}{243,202,120}
\definecolor{hidden-yellow}{RGB}{242,244,193}

\usepackage{amssymb}
\usepackage{pifont}
\begin{document}
%
\title{UniMS-RAG: A Unified Multi-source Retrieval-Augmented Generation \\ for Personalized Dialogue Systems}
%
%
%

\author{Hongru Wang, Wenyu Huang, Yang Deng, Rui Wang, \\ Zezhong Wang, Yufei Wang, Fei Mi, Jeff Z. Pan, Kam-Fai Wong 

\thanks{Hongru Wang, Rui Wang, Zezhong Wang, and Kam-Fai Wong are with the Department
of Systems Engineering and Engineering Management, The Chinese University of Hong Kong. e-mail: ({hrwang, kfwong}@se.cuhk.edu.hk).}
\thanks{Wenyu Huang and Jeff Z. Pan is with EdinburghNLP, The University of Edinburgh, United Kingdom.}%
}

\maketitle

\begin{abstract}
Large Language Models (LLMs) have shown exceptional capabilities in many natural language understanding and generation tasks. However, the personalization issue still remains a much-coveted property, especially when it comes to the multiple sources involved in the dialogue system. To better plan and incorporate the use of multiple sources in generating personalized response, we firstly decompose it into three sub-tasks: Knowledge Source Selection, Knowledge Retrieval, and Response Generation. We then propose a novel Unified Multi-Source Retrieval-Augmented Generation system (UniMS-RAG) Specifically, we unify these three sub-tasks with different formulations into the same sequence-to-sequence paradigm during the training, to adaptively retrieve evidences and evaluate the relevance on-demand using special tokens, called acting tokens and evaluation tokens. Enabling language models to generate acting tokens facilitates interaction with various knowledge sources, allowing them to adapt their behavior to diverse task requirements. Meanwhile, evaluation tokens gauge the relevance score between the dialogue context and the retrieved evidence. In addition, we carefully design a self-refinement mechanism to iteratively refine the generated response considering 1) the consistency scores between the generated response and retrieved evidence; and 2) the relevance scores. Experiments on two personalized datasets (DuLeMon and KBP) show that UniMS-RAG achieves better performance than previous strong baselines on the knowledge source selection and response generation task with itself as a retriever in a unified manner, and achieves new state-of-the-art when using more advanced external retriever. Extensive analyses and discussions are provided for shedding some new perspectives for personalized dialogue systems.
\end{abstract}


%
\IEEEpeerreviewmaketitle

\section{Introduction}

\IEEEPARstart{T}{the} emergence of large language models (LLMs) has revolutionized the field of natural language processing, including many downstream understanding and generation tasks \citep{bang2023multitask}. While these models have undeniably advanced the state of the art in various applications, they also introduce new challenges, particularly the factual error \citep{mallen-etal-2023-trust,xue2023improving} and personalization issues \citep{salemi2023lamp, wang2023cuecot} in the realm of dialogue systems. To alleviate this, Retrieval-Augmented Generation (RAG) methods are usually adopted to retrieve relevant passages, aiming to enrich the semantic information of the dialogue context, resulting in up-to-date, factual and personalized responses. Furthermore, the latest Self-RAG \citep{asai2023selfrag} considers the cases that require external knowledge and the cases that do not at the same time, achieving a better trade-off between effectiveness and efficiency. In a world inundated with vast amounts of information, retrieval-augmented dialogue systems play a crucial role in ensuring that automated conversations are not only linguistically proficient but also factually reliable and contextually aware. 

\begin{figure}[t]
  \centering
  \includegraphics[trim={3cm 8cm 4cm 10cm}, clip, width=1.0\linewidth]{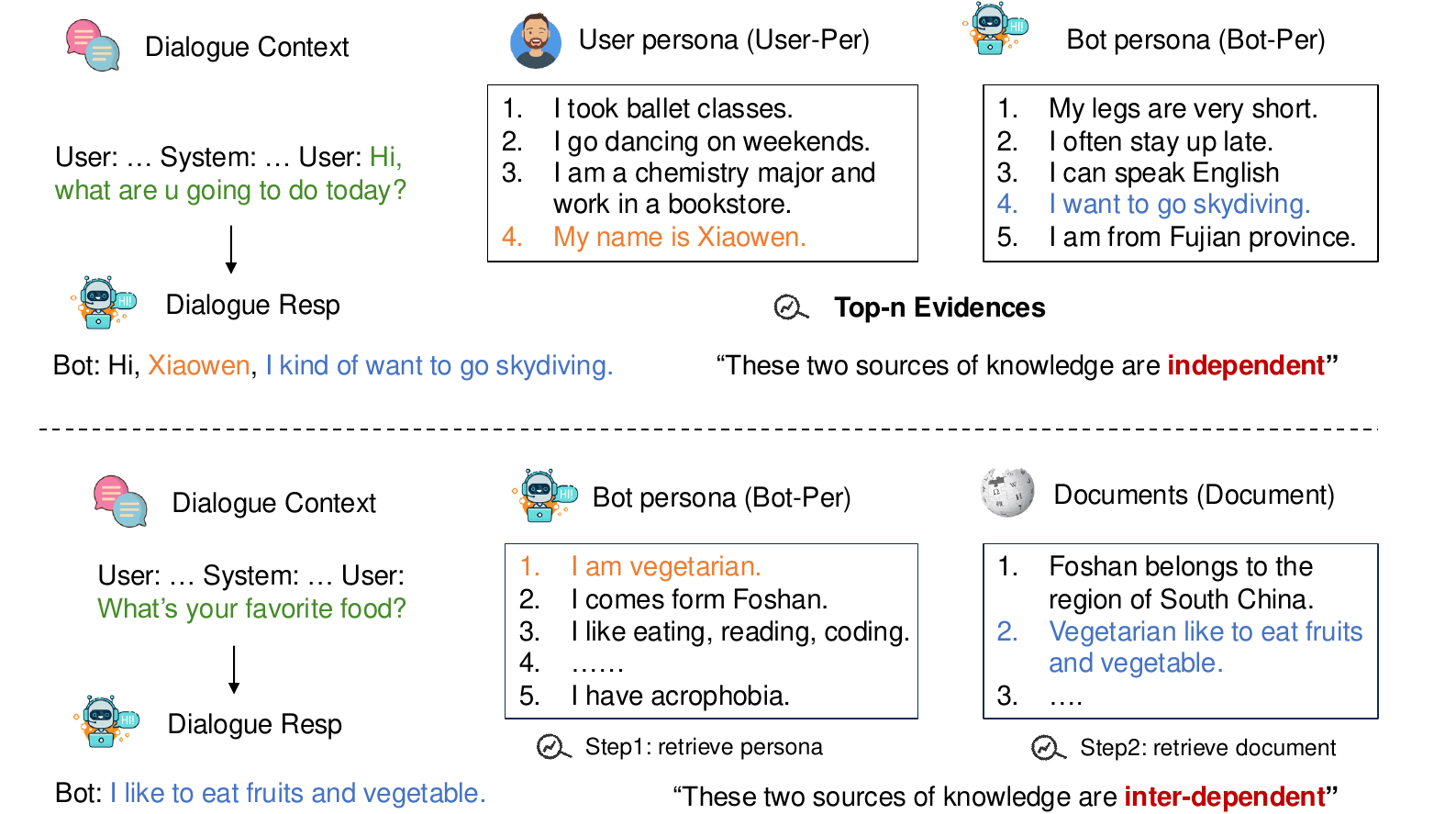}
  \caption{Two typical examples of multi-source personalized knowledge-grounded dialogues: \textbf{upper)}: An example from DuLeMon \citep{dulemon}; and \textbf{bottom)}: An example from KBP \citep{wang2023large}. We use same color to indicate the response and corresponding grounded knowledge. We skip the dialogue context for simplicity.}
\label{intro_examples}
\end{figure}

There are two limitations. On the one hand, most of the existing knowledge-grounded dialogue systems either focus on a single source of knowledge or indiscriminately incorporate all sources of knowledge. In detail, previous methods usually interact with different external sources of knowledge to engage the user and provide human-like conversational experience, such as system persona source to maintain consistency in responses \citep{dulemon}, user memory or profile source for personalized responses \citep{madotto-etal-2019-personalizing, salemi2023lamp}, Wikipedia \citep{dinan2019wizard} or the internet \citep{komeili-etal-2022-internet} source to ensure the provision of up-to-date information. Yet, they focus on individual sources or tasks in isolation, overlooking the complexity and the potential existence of multiple sources of knowledge in real-world scenarios, as shown in Figure~\ref{intro_examples}. In this way, the dialogue system needs to decide which source to use (or not use) dynamically, in order to provide more personalized and helpful responses. Furthermore, the inter-dependency relationship between these different sources of knowledge needs to be considered to guide better response generation \citep{wang2023large}. For example, as shown in bottom part of Figure~\ref{intro_examples}, the dialogue system needs to locate the specific persona config first according to the dialogue context, and then find relevant dependent documents to answer the user inquiry. On the other hand, previous work either independently train the retriever and reader \citep{wang2023large, ram2023incontext}, resulting in sub-optimal performance and distribution shift between retriever and reader; or design complex architecture to optimize them at the same time \citep{realm,rubin2023longrange}, which is infeasible and inefficient at the era of large language models due to unaffordable computing cost. 

To tackle the aforementioned challenges, we first decompose the problem of \textbf{personalized knowledge-grounded dialogue response generation task (PerDS)} into three different tasks:

\begin{itemize}[leftmargin=*]
    \item \textit{Knowledge Source Selection (planner)} aims to plan the source call order to guide the knowledge retrieval, considering both the independent or inter-dependent relationships within different sources.
    \item \textit{Knowledge Retrieval (retriever)} sequentially retrieves \textit{top-n} evidence from external sources according to the decisions in the last step. 
    \item \textit{Response Generation (reader)} produces knowledge-grounded natural language responses to users according to original dialogue context and retrieved evidence.
\end{itemize}

Then we design a novel framework, \textbf{Unified Multi-Source Retrieval-Augmented Dialogue System (UniMS-RAG)}, that unifies three tasks in the above using the same large language models in a \textbf{sequence-to-sequence (Seq2Seq)} manner. In specific, motivated by recent work of assigning different tokens with different roles \citep{wang2023large, asai2023selfrag}, we carefully introduce two types of tokens: 1) \textbf{acting tokens} to decide the next action (e.g., which source to use), aiming to call different source on-demand instead of incorporating all of them; and 2) and \textbf{evaluation tokens} to evaluate the relevance score between dialogue context and retrieved evidence (e.g., the similarity score), in order to force the model attention on more relevant evidence while overlooking noisy ones. Thus we can reformulate the above three tasks as token prediction tasks by generating acting tokens (planner), evaluation tokens (retriever) or original tokens in the vocabulary (reader) during the training. We randomly shuffle the order of retrieved evidences to prevent model learning a shortcut by attentions on evidences appears in specific positions. To further enhance the quality of generated responses, we incorporate a self-refinement process during the inference stage. This involves reassessing the generated responses by leveraging feedback from evaluation tokens and ensuring consistency between the provided evidence and the responses. To sum up, the contributions are summarized as follows:

\begin{itemize}[leftmargin=*]
    \item We formally propose a multi-source personalized knowledge-grounded dialogue tasks, consisting of three different sub-tasks: \textit{knowledge source selection}, \textit{knowledge retrieval} and \textit{final response generation}. Our analysis provides detailed insights into the limitations and proficiencies of current LLMs across these sub-tasks.
    
    \item We propose a novel method, namely, UniMS-RAG, that tackles all sub-tasks in PerDS with a unified model. To the best of our knowledge, it is the first attempt to utilize LLMs as the planner, retriever and reader at the same time.

    \item We investigate the different strategies to get the soft label of evaluation tokens during the training stage, including prompting LLMs or using an independent fine-tuned retriever (a.k.a, \textit{classification-based} and \textit{prompting-based methods}). Furthermore, we propose a self-refinement mechanism to re-generate the response using updated evidence according to its relevance with the dialogue context and previously generated response.
    
    \item Experimental results on two PerDS benchmark datasets show that UniMS-RAG outperforms previous strong baselines, and achieves state-of-the-art performance with external more effective retriever, resulting in more personalized and factual responses. Extensive analyses provide some new insights into the future of multi-source retrieval-augmented generation tasks.
\end{itemize}

\section{Related Work}

\subsection{Personalized Dialogue System}
To build a personalized dialogue agent, Zhang et al. \cite{personachat} firstly investigated this task with a new dataset Persona-Chat, where a pre-defined persona set is a form of multiple sentences of textual description. Lots of works follow this setting and have taken mutual persona perception \citep{liu-etal-2020-impress, cosplay}, persona-sparse scenario \citep{bob, welch-etal-2022-leveraging}, long-term persona memory \citep{dulemon}, persona extending \citep{persona_extending} and persona order bias \citep{chen-etal-2023-towards-robust} into consideration. Although some of them complement the insufficient semantics in short persona descriptions by further utilizing an external commonsense knowledge base to extend existing persona sets \citep{hiking, persona_extending}, they still fall into the conventional framework coupling the knowledge selection with the response generation \cite{wu-etal-2022-ksam}, rendering it infeasible to handle various sources of knowledge. There have also been works showing that the combination of different knowledge sources such as persona descriptions and Wikipedia can further improve the overall performance~\cite{focus,wu-etal-2021-better,wu-etal-2022-section}. However, they still fail to capture possible dependency between knowledge sources. In their framework, knowledge is not used as the role to assist persona-consistent response generation, but as an additional resource to generate a more informative response \citep{wow, xue2023improving} or select a suitable persona \cite{focus,fu-etal-2022-thousand}. Furthermore, most existing works overlook the possibilities that the response does not require the involvement of persona descriptions by simply concatenating all personas with the dialogue context to generate the final response \citep{hiking, cosplay}.

\subsection{Knowledge-grounded Dialogue System}

How to interact with different external sources plays a key role in dialogue systems to retrieve corresponding knowledge, resulting in more helpful, personalized, trustworthy responses \citep{10.1145/3383123, wang2023survey}. Specifically, most of previous methods rely on different external sources of knowledge to engage the user and improve the conversational experience, including but not limited to \textit{system persona} to maintain consistency in responses \citep{wang2023large}, \textit{user memory or profile} for personalized interactions \citep{xu-etal-2022-long, 10.1145/3317612}, and access to \textit{sources like Wikipedia or the internet} to ensure the provision of up-to-date information \citep{dukenet, nakano2022webgpt}. However, most of them focus on individual sources or tasks in isolation, overlooking the complexity and the potential existence of multiple sources in practice. There is a latest work named TPE which regards different knowledge sources as \textit{conceptual tools} and proposes a multi-persona collaboration framework to model the decision-making process of the call order for multiple knowledge sources \citep{wang2023tpe}. We differ in exploring the capability of LLMs to be planner, retriever, reader in a unified manner.

\subsection{Retrieval-augmented Generation}

Retrieval-augmented Generation (RAG) has been considered as an effective method to overcome several limitations of LLMs, such as hallucinations \citep{shuster-etal-2021-retrieval-augmentation}, factuality \citep{factuality_survey} and long-term memory \citep{dulemon}. Usually, an external retriever is first used to retrieve relevant textual knowledge from one specific knowledge source (e.g., Wikipedia), then the reader takes the relevant textual knowledge as external context for generating knowledge-grounded response \citep{rag}. Most of previous works try to optimize the retriever and reader independently \citep{gao2024retrievalaugmented}. During the initial phases, people use sparse retriever, such as BM25 \citep{bm25}, to make relevance decisions and retrieve corresponding evidence. However, sparse approaches fall short in extracting the semantic features inherent in text content \citep{retriever_survey}. To overcome this issue, researchers have proposed language model-based dense retrieval methods by encoding documents and queries as dense vectors, which effectively represent the semantic features of text content \citep{dpr, 10.1145/3570724, 10.1145/3596512}. For example, DPR \citep{dpr} uses two pre-trained language models to encode documents and queries separately, allowing for a more nuanced understanding of the content. More recently, there are a handful of works exploring the performance of LLMs as retriever \citep{zhu2023large, sun-etal-2023-chatgpt,shen2023large, ma2023zeroshot}. In detail, Shen et al. \cite{shen2023large} firstly prove that LLMs can be used as strong zero-shot retriever on several benchmark datasets, while Ma et al. \cite{ma2023zeroshot} propose Listwise Reranker with a Large Language Model (LRL), which achieves strong reranking effectiveness without using any task-specific training data. Distinguishing from previous works, we finetune LLM itself to learn a joint distribution of dialogue and evidence using similarity feedbacks from current most powerful LLMs, such as ChatGPT and GPT-4.



\section{Problem Definition}
To consider the responses which require external knowledge and those which do not in practice, we provide a unified definition to unify the retrieval-augmented dialogue system and non-retrieval dialogue system, following Wang et al. \citep{wang2023large}. Let $\boldsymbol{C}_t =\{u_1, s_1, u_2, s_2, ..., u_t\}$ denotes the dialogue context at the current conversation turn $t$, and there are different knowledge sources $\boldsymbol{K}=\{K_1, K_2, ..., K_i\}$, where $K_i = \{k_i^1, k_i^2, ..., k_i^j\}$ indicates the $i_{th}$ source's name of $\boldsymbol{K}$. $k_i^j$ denotes the $j_{th}$ knowledge in natural language from $K_i$. The goal of retrieval-augmented dialogue system\footnote{In this context, a non-retrieval dialogue system is viewed as a specific type within retrieval-augmented dialogue systems, distinguished by the absence of a knowledge source, denoted as \texttt{NULL}.} is to select suitable knowledge sources, and then generate the helpful, informative and personalized response, depending on which source is chosen \citep{wang2023survey}. Thus, the problem of \textbf{PerDS} can be decomposed into the following three tasks:

\begin{itemize}[leftmargin=*]

    \item \textit{\textbf{Knowledge Source Selection.}} At each turn $t$, given the dialogue context $C_t$ and different knowledge sources $\boldsymbol{K}$, PerDS first select suitable sources denoted as $S \in \{\boldsymbol{K}, \texttt{NULL}\}$. It's important to highlight that $S$ may consist of multiple sources or be NULL. Notably, there are no constraints imposed on the relationships between different sources within $S$. They can either be independent or interdependent.
    
    \item \textit{\textbf{Knowledge Retrieval.}} The second task is to retrieve \textit{top-n} corresponding evidence $\boldsymbol{E} = \{e_1, e_2, ..., e_n\}$ for each selected source if there is. We simply skip this step once the PerDS determine there is no need to call external knowledge.

    \item \textit{\textbf{Response Generation.}} The final task is to generate a proper response $s_t$ concerning the dialogue context $\boldsymbol{C_t}$ and necessary evidences $\boldsymbol{E}$ from different external knowledge sources $S$ if there is. The generated response is expected to ground on these evidences, being personalized, informative, up-to-date and helpful according to the distinctions across different sources.
    
\end{itemize}

\section{Method}
\begin{figure}[t]
  \centering
  \includegraphics[trim={0cm, 6cm, 0cm, 11cm}, width=\linewidth]{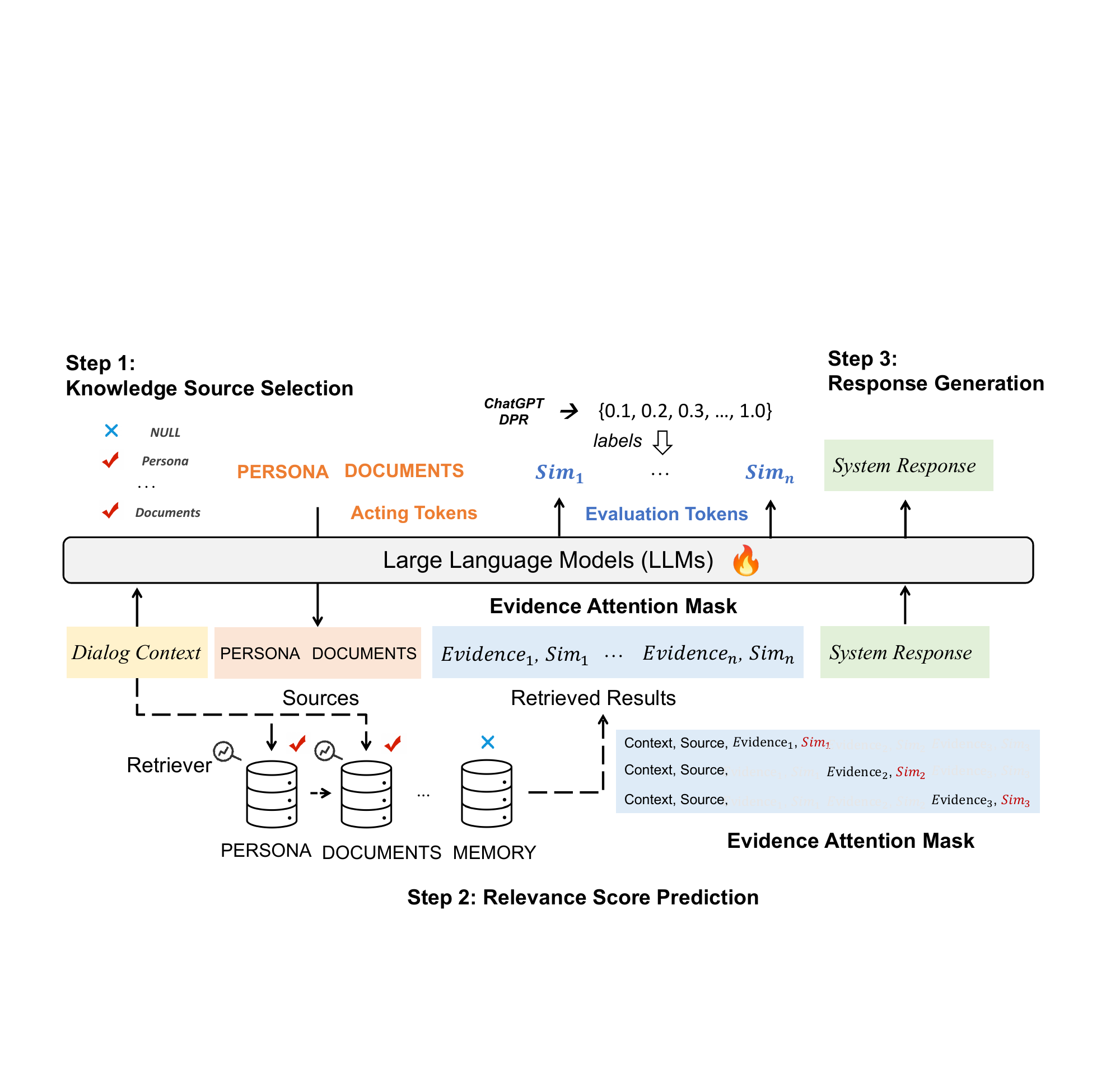}
  \caption{Our proposed method UniMS-RAG, where three optimization tasks are carefully designed: 1) Knowledge Source Selection; 2) Relevance Score Prediction; and 3) Response Generation. We use orange to indicate \textcolor{orange}{acting tokens} and blue to indicate \textcolor{blue}{evaluation tokens}. It is worth noting we have all labels during training to optimize these three sub-tasks in a teacher-forcing way.}
  \label{unims_rag}
\end{figure}

In this section, we first describe the framework that reformulates each task in PerDS into the unified Seq2Seq paradigm, and then introduce the joint-training strategies for the retriever and reader modules and the inference strategy to re-evaluate the quality of response, respectively. The overview and examples of the input and output sequences for UniMS-RAG are illustrated in Figure~\ref{unims_rag}.

\subsection{UniMS-RAG}
Inspired by previous work \citep{wang2023large}, we propose an innovative methodology termed \textbf{U}niMS-RAG, where $U$ signifies the unification of the training process for planner, retriever and reader, as well as the integration of diverse tasks into a singular comprehensive framework. Recognizing the potential of large language models (LLMs) in orchestrating the utilization of external sources of knowledge, as indicated by recent works, UniMS-RAG extends the capabilities of LLMs to seamlessly connect disparate sources within the context of personalized knowledge-grounded dialogues. This integration streamlines the traditionally separated tasks of retriever and reader training, enabling to adaptively retrieve evidences and evaluate the relevance scores in a unified manner.

To address the interactions between different subtasks, as illustrated in Figure~\ref{unims_rag}, the whole response generation can be modelled into three steps in UniMS-RAG: 1)~\textbf{Planning}, to make a series of decisions about whether to use a specific knowledge source given relationship descriptions between different sources; 2)~\textbf{Retrieval}, to retrieve \textit{top-n} results from external databases according to the decisions; 3)~\textbf{Generation}, to incorporate all retrieved knowledge (if required) into the final response generation. Taking advantage of the decoupling of these different tasks, the UniMS-RAG framework exhibits versatility and scalability in its applicability to various retrieval-augmented response generation tasks. For example, it can be achieved through targeted modifications to the \textit{Planning} step. By configuring decisions within this phase, the model can seamlessly accommodate different retrieval-augmented tasks without necessitating extensive adjustments to other components.

\subsubsection{Planning} 

First of all, to incorporate the cases which does not require any sources of external knowledge, we define several additional indicate tokens, corresponding to different sources, including the \texttt{NULL} token which signifies that there is no requirement for any external knowledge as shown in Table~\ref{tab:act_tokens}:
\begin{equation}
    \mathcal{M}: \boldsymbol{c} \rightarrow \texttt{NULL},
\end{equation}
where $\mathcal{M}$ is parameterized by LLMs. Secondly, there are two different situations between these multiple knowledge sources: 1) the $K_1, K_2, ..., K_i$ in $\boldsymbol{K}$ is independent, which means there is no interdependent relationship between them; and 2) Some or even all of they are not independent, for example, the results obtained from $K_1$ may be contingent on the outcomes derived from $K_2$ and potentially other sources. These two situations cater to different applications. In the first scenario, the independence between knowledge sources offers practical utility for tasks where autonomy and isolation of information are paramount, such as user persona and system persona, which are two independent knowledge sources. On the other hand, in the second scenario, where interdependencies appear, the model accommodates tasks that demand a nuanced consideration of the relationships between different knowledge sources. For example, dependencies between user memory or persona and document introduce a layer of complexity, reflecting real-world scenarios where the fact that age, hobby, education, and life experience of the user have a major effect on his or her personal preference over external document knowledge \citep{fu-etal-2022-thousand}. 

\begin{table*}[]
    \centering
    \caption{Three types of special tokens used in UniMS-RAG, in which the former two stands for acting tokens and evaluation tokens, respectively. Each type uses several tokens to represent its output values.}
    \begin{adjustbox}{width=0.8\textwidth}
    \begin{tabular}{c|c|c|c}
    \toprule
     Token Type  & Input & Output & Definitions \\
     \midrule
     Sources   & $\boldsymbol{C}$ & $\{\texttt{NULL}, K_1, K_2, K_3, ... K_n\}$ & Decides which source to retrieve or not retrieve. \\
     \midrule
     Similarity & $\boldsymbol{C}, e_i$ & $\{0.0, 0.1, 0.2, ..., 1.0 \}$ & $e_i$ is a useful for current dialogue context.\\
     \midrule
     Indicator & - & $\{\texttt{[SOURCE]}, \texttt{[EOS]}, \texttt{[EVIDENCE]}, \texttt{[EOE]}\}$ & Indicates the start and end position of different parts. \\
     \bottomrule
    \end{tabular}
    \end{adjustbox}
    \label{tab:act_tokens}
\end{table*}

In order to handle both independent and interdependent knowledge sources, thereby extending its applicability across a spectrum of use cases with varying degrees of complexity, the goal of the planning step is to make a series of decisions to decide whether or not the corresponding source of knowledge is required and determine their call order if needed. Since the dependency relationship is previously known, we only need to make sure that a certain knowledge source is called after the sources it depends on. Thus, we formulate this task as sequence-to-sequence generation by directly outputting required sources in execution order as follows:
\begin{equation}
    \mathcal{M}: \boldsymbol{c} \rightarrow K_i, K_j, ..., K_n,
\end{equation}

Then we strictly follow the outputed order to retrieve evidences from corresponding source of knowledge. To offer the flexibility and scalability to plug in an arbitrary number of sources, we can add $K_i, ..., K_n$ and \texttt{NULL} as special tokens into the vocabulary of LLMs as special tokens, and expand the set of tokens on the fly following Hao et al. \citep{hao2023toolkengpt}. Besides that, we add other special tokens to indicate the different parts of the input, \textit{i.e.,} \texttt{[SOURCE]} and \texttt{[EOS]} to indicate the start and end positions of sources. In this way, LLM can model the dependency between different sources and learn when and how to call certain sources.

\subsubsection{Retrieval}

According to the output of the last \textit{planning} step, there are two cases in this step: (1) the response does not need any external sources of knowledge, and the dialogue system can skip this step; and (2) the response needs multiple sources of knowledge, and it strictly follows the output source order to retrieve \textit{top-n} related evidence $k_i^{*}$ for the $i_{th}$ source of knowledge according to the dialogue context $\boldsymbol{c}$. If there is a dependency here, it will use preceding retrieved results $k_{j}^*$ in the planned execution order as a filter. Specifically, assuming the output order is \texttt{PERSONA}, \texttt{DOCUMENTS} in the planning step for a persona-consistent dialogue system, we first retrieve \textit{top-1} result $p^*$ from \texttt{PERSONA}, and then we retrieve $k^*$ from \texttt{DOCUMENTS} according to $\boldsymbol{c}$ and $p^*$. If there is no dependency, it simply retrieves corresponding evidences source by source (as shown in Figure~\ref{intro_examples}).

To joint train the retriever and reader in a unified manner, we first reformulate the traditional classification task (relevant or irrelevant) for retrieval into a generation task. Then we define several similarity tokens as shown in Table~\ref{tab:act_tokens}, ranging from 0 to 1 with the internal as 0.1. In this way, we enforce the UniMS-RAG to predict the similarity score by generating corresponding tokens. To get the supervised label of similarity scores, there are two different ways: 1) the similarity scores are generated by independent trained retriever such as DPR \citep{dpr}; 2) the similarity scores are generated by prompting LLMs \citep{zhu2023large}. The details can be found in \S~\ref{retriever_training}. Furthermore, we design a \textbf{evidence attention mask} mechanism to mask unrelated evidence when predicting the similarity score for current evidence, aiming to reduce unnecessary noises in the context as shown in Figure~\ref{unims_rag}. 
\begin{equation}
\label{unims_rag_retrieval}
    \mathcal{M}: \boldsymbol{c}, K_i, e_j \rightarrow sim \in \{0.1, 0.2, ..., 1.0 \},
\end{equation}

In this way, UniMS-RAG can be used to evaluate the relevance between the dialogue context and retrieved evidences after training, serving as a retriever itself. During the inference, we can use it to retrieve top-n evidences $\{e_1, ..., e_n\}$ from corresponding sources of knowledge according to Eq.~\ref{unims_rag_retrieval}.

\subsubsection{Generation}

We concatenate all preceding results, including the names of sources, retrieved evidences and corresponding similarity scores, all together with the dialogue context $\boldsymbol{c}$ to generate the response:
\begin{equation}
    \mathcal{M}: \textit{Inp} \rightarrow s_t,
\end{equation}
where $\textit{Inp} = \{ \boldsymbol{C}_t \quad \texttt{[SOURCE]} K_i, ..., K_n \texttt{[EOS]} \quad \texttt{[EVIDENCE]} \\ k_i^j \texttt{[EOE]} [Sim_1] ,..., \texttt{[EVIDENCE]} k_n^m \texttt{[EOE]} [Sim_n] \}$. We use \texttt{[EVIDENCE]} and \texttt{[EOE]} to represent the start and end positions of the retrieved evidences. The $[Sim_i]$ stands for the similarity score of $i_{th}$ evidence, calculated using the retriever (\S\ref{retriever_training}).

\subsection{General Framework}
\label{retriever_training}

\begin{figure}[t]
    \includegraphics[trim={8cm 9cm 10cm 5cm}, clip, width=0.9\linewidth]{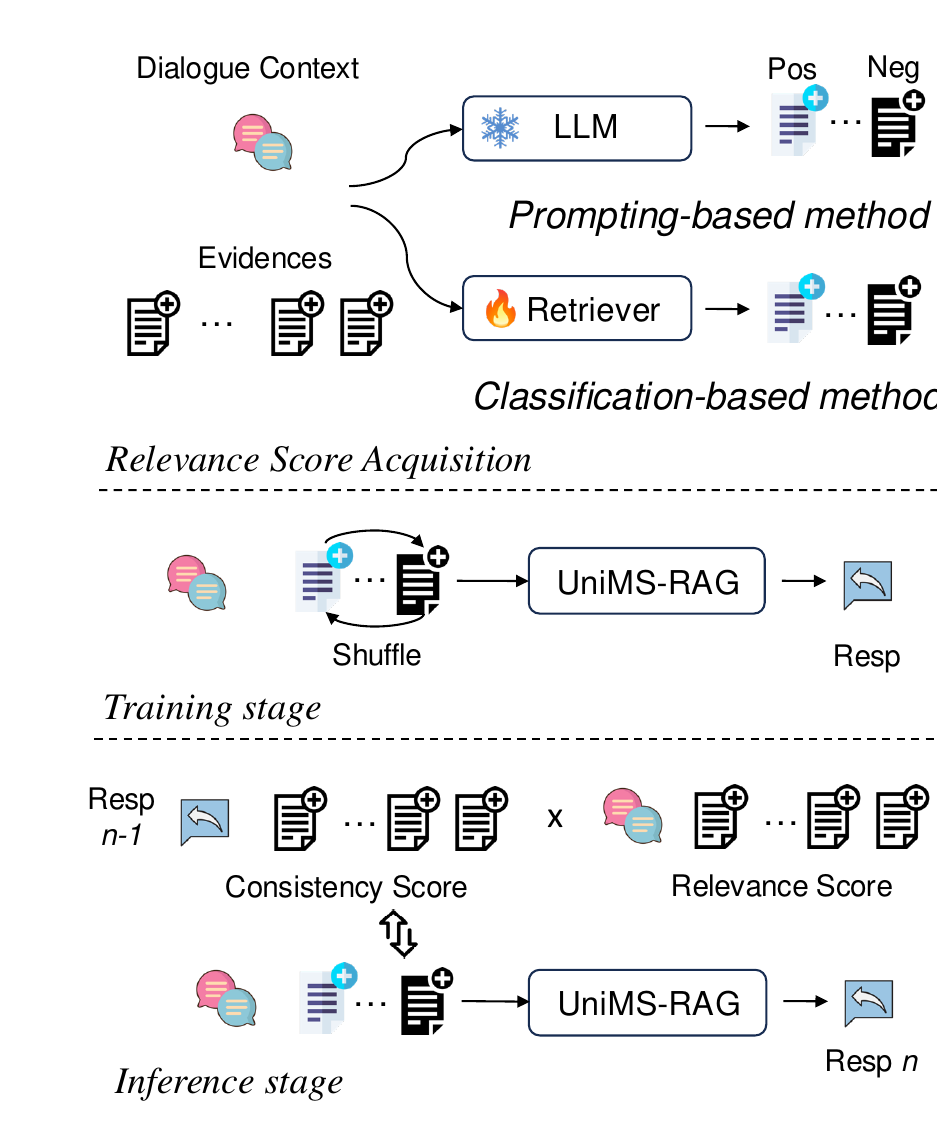}
    \caption{The general framework to utilize UniMS-RAG, including 1) relevance score acquisition; 2) training stage; and 3) inference stage.}
    \label{fig:general_framework}
\end{figure}

Besides the core model, we also value the general framework such as different ways to collect data, unique training and inference strategies. Thus, we first introduce different ways to collect relevance score labels, and then present the training and inference design (Figure~\ref{fig:general_framework}).

\subsubsection{Relevance Scores Acquisition}
\label{joint-training_retriever}

There are two different methods to get the similarity labels: 1) \textit{prompt-based method.} which directly prompt the LLMs to assign the similarity scores given dialogue context and evidences \citep{sun-etal-2023-chatgpt, ma2023zeroshot}; 2) \textit{classification-based method.} which requires a fine-tuned retriever such as DPR to calculate the similarity score given the same input. In latter method, the evidence is feed one by one with the dialogue context unchanged.

\begin{figure}[t]
    \includegraphics[trim={0cm 3cm 0cm 2cm}, clip, width=\linewidth]{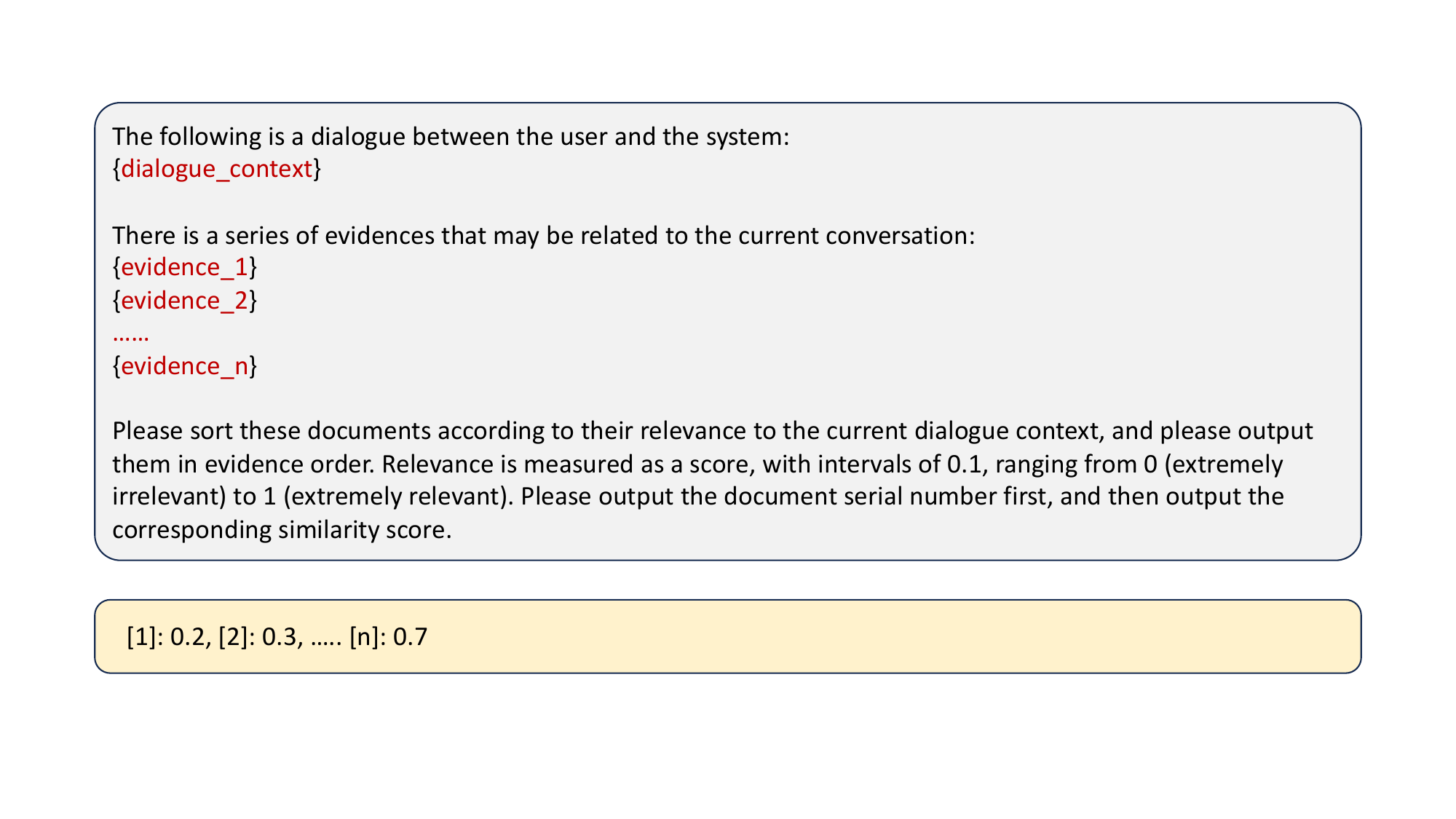}
    \caption{The instructions for zero-shot retriever used to predict similarity score using off-the-shelf LLMs. The grey and yellow blocks indicate the inputs and outputs of the model.}
    \label{llm_sim_prompt}
\end{figure}

\noindent \textbf{Prompt-based method.} We can utilize some off-the-shelf methods to get the similarity score between the dialogue context and the evidence, including some sparse retriever such as TF-IDF \citep{sparck1972statistical} and BM25 \citep{bm25}, or simply hard label by assigning a similarity score to the evidence based on whether it is used (set score as 1) or unused (set score as 0). Inspired by recent studies which simply using LLMs as search engine to predict the similarity relationship between query and evidence \citep{ma2023zeroshot}, we choose to prompt the LLMs (i.e, ChatGPT) to predict the similarity score by feeding a dialogue context and several evidences in a zero-shot manner. The prompts are shown in Figure~\ref{llm_sim_prompt}.

\noindent \textbf{Classification-based method.} Following previous works \citep{dpr}, we can train an off-the-shelf retriever using the ground-truth labels. Specifically, we first build our own finetuning dataset by regarding \textit{(context, related\_evidence)} as the positive and \textit{(context, unrelated\_evidence)} as the negatives. The positive and negative samples are annotated in the original dialogue dataset, i.e., which persona or documents is used to generate the response. Then we apply the following negative log likelihood (NLL) loss to learn the retrieval-oriented representations following \citep{fs-cdr}:

\begin{equation}
\mathcal{L}_{retriever} = - log \frac{e^{sim(C_i, d^{+})}}{ e^{sim(C_i, d^{+})} + \sum_{j\in \mathcal{B^-}}{e^{sim(C_i,d^-)} }},
\end{equation}
\begin{equation}
    sim(C, d) = cos (C, d))
\end{equation}

where \textit{sim} is a similarity function, $\mathcal{B}$ is a mini-batch of examples, $d^+$ and $d^-$ are positive evidence and negative evidence for $i_{th}$ dialogue context $C_i$. Once there are no negatives or the negative is the same as the positive, we directly use randomly sampled sample from other session. By including mini-batch negative candidates in the training data, the model is forced to learn to identify the subtle and key information required for the current turn instead of useless or redundant ones. After training, we can use the fine-tuned DPR to provide relevance labels to train the UniMS-RAG.

To avoid computing on the fly during training, we pre-compute relevance scores for all training and validation samples before beginning the model training process. During the inference stage, three types of retrievers can be employed: 1) DPR, 2) LLMs, and 3) UniMS-RAG. It is worth noting that UniMS-RAG can be considered a distillation derived from either DPR or LLMs.

\subsubsection{Training Stage} 

We shuffle the order of evidence during the training. The formation of the input in this way has three advantages. Firstly, the name of the sources indicates the type of results retrieved, which provides more signals to the LLMs. Secondly, it enforces the LLMs to capture the important evidence since it can appear in any position and denoise unrelated evidence. Thirdly, it allows us to train the large language model in a multi-task manner using teacher forcing as introduced in next section. The whole training objective of UniMS-RAG can be defined as follows:

\begin{equation}
    \mathcal{L} = \mathcal{L}_{source} + \mathcal{L}_{sim} + \mathcal{L}_{response}
\end{equation}

Where the $L_{source}, L_{sim}, L_{response}$ indicate planning loss, relevance predication loss and the final response loss. In other words, we only calculate the loss at acting, evaluation and response tokens. By assigning different tokens different roles, our UniMS-RAG can serve as the planner (determine whether or not use external sources of knowledge and then use which source), retriever, and reader in a unified manner.

\subsubsection{Inference Stage}

In this section, we introduce how to refine the generated responses, considering the similarity score between dialogue context and retrieved evidences and consistency score between retrieved evidences and generated responses. Algorithm~\ref{alg:self-refinement} shows the details of the whole processing. Specifically, we first calculate the consistency score between each retrieved evidence ($e_i$) and current generated response ($r$) using a fine-tuned natural language inference model (see \S~\ref{implementaion_details}). Next, we determine the overall score for each evidence by multiplying its similarity score with the consistency score. This combined score serves as an indicator of the evidence's quality, taking into account its relevance to the dialogue context and its alignment with the generated response. In this way, we can adjust the evidences based on their determined quality and a predefined update number, resulting in the creation of a new list of evidences. This process ensures that the most relevant and high-quality pieces of evidence are given priority, contributing to the refinement of the overall generated responses. Ultimately, we can re-generate the responses $r_{new}$ using a new list of evidences, providing the dialogue context. Moreover, we can iteratively apply the self-refinement algorithm by incorporating the newly generated responses, denoted as $r_{new}$, and the updated list of evidences. We provide a thorough analysis at \S~\ref{infer_feedback}, outlining the rationale behind the selection of update numbers and the procedural steps involved in refining the responses. 

\begin{algorithm}[t]
    \caption{Inference with Self-refinement} 
    \label{alg:self-refinement}
    \begin{algorithmic}[1]
    \algsetup{linenosize=\small}
    \REQUIRE Dialogue Context $C$, UniMS-RAG model $M$, NLI model $N$, Retriever $R$, Generated Response $r$, Sources of Knowledge $\boldsymbol{K}= \{ K_1, K_2, ..., K_m \}$, Retrieved Evidences $E = \{e_1, e_2, ..., e_n \}$, Similarity Scores $S_{ce} = \{s_1, s_2, ..., s_n\}$, Update Number $\alpha$
    
    \ENSURE Refined Response $r_{new}$.

    \STATE $S_{nli}$ = [] // store all nli scores
    \FOR{$j=1,2,...,n$}
    \STATE $nli_j$ = $N (e_j, r)$ // get the nli consistency score
    \STATE $S_{nli}.append(nli_j)$
    \ENDFOR

    \STATE S = $S_{nli}$ * $S_{ce}$ // element-wise product inspired by $p(r|c) = p(e|c)p(r|c,e)$ \\ 
    \STATE S = sorted (S) // sort the above scores
    \STATE $E_{update}$ = Find ($\alpha$, S, $E$) // find the evidences need to be updated according to the sorted scores
    \STATE $E = E \setminus E_{update}$ // pop all evidences need to be updated from current one
    \WHILE{$E_{update} \neq \emptyset$}
    \STATE Pop a evidence $e_{i}$ from $E_{update}$, then $E_{update} = E_{update} \setminus \{e_i\}$
    \STATE $e_{new}$ = R (C, K) // retrieve next novel evidence from corresponding source of knowledge according to the dialogue context
    \STATE $E = E \cup \{e_{new}\}$
   
    \ENDWHILE
    \STATE $r_{new}= M ( C, E )$.
    \RETURN $r_{new}$
    \end{algorithmic}
\end{algorithm}

\section{Experimental Setups}

\subsection{Research Questions}

The empirical analysis targets the following research questions:

\begin{itemize}[leftmargin=*]
    \item \textbf{RQ1:} Can large language models serve as a \textbf{planner} to determine whether or not require knowledge, which source of knowledge to call, and when to call?
    \item \textbf{RQ2:} Can large language models serve as a \textbf{retriever} to retrieve highly related evidence from corresponding sources of knowledge?
    \item \textbf{RQ3:} Can large language models serve as a \textbf{reader} to capture related information in the context and incorporate them into the final response generation?
\end{itemize}

\subsection{Datasets}

\begin{table}[]
    \centering
    \caption{Statistics of DuLeMon and KBP Datasets}
    \begin{adjustbox}{max width=0.48\textwidth}
    \begin{tabular}{lrr}
    \toprule
     Dataset  & DuLeMon & KBP  \\
     \hline
     \#Dialogues & 3011 & 2477 \\
     Train/Val/Test & 19437/2407/2416 &  9821/1227/1229 \\
     \#Utterances & 24554  & 48522 \\
     Source Types & \texttt{NULL}, \texttt{User-Per}, \texttt{Bot-Per} & \texttt{NULL}, \texttt{Persona}, \texttt{Document} \\
     Resp w/ source & $\approx$ 49\%  & $\approx$ 85\% \\
     \bottomrule 
    \end{tabular}
    \end{adjustbox}
    \label{data_stat}
\end{table}

We consider two different situations between different knowledge sources, corresponding to two publicly available personalized dialogue datasets: DuLeMon
\cite{dulemon} and KBP 
\citep{wang2023large}. The dataset statistics are presented in Table~\ref{data_stat}. We adopt the same train/val/test split in the original datasets.

\begin{itemize}[leftmargin=*]
    \item \textbf{DuLeMon} \citep{dulemon} is the latest open-domain dialogue dataset with long-term persona memory in which a response is grounded on persona information that occurred in historical sessions, leading to better dialogue engagement. There are two versions of DuLemon: \textit{Self} and \textit{Both}, where the persona information comes from only \texttt{self} side (user side) or \texttt{both} side (both user and chatbot side). We choose \textit{Both} versions to consider the independent relationship between these two sources of persona information: User side (\texttt{User-Per}) and Chatbot Side (\texttt{Bot-Per}).

    \item \textbf{Knowledge Behind Persona (a.k.a., KBP)} \citep{wang2023large} is a dialogue dataset, in which the response is grounded on both the persona (\texttt{Persona}) and its corresponding implicit knowledge (\texttt{Document}). We choose it to consider the interdependent relationship between different sources of knowledge. It is worth noting that both of these two datasets contain cases which do not require any external source of knowledge, denoting as \texttt{NULL}.
\end{itemize}

\subsection{Baselines and Evaluation Metrics}
\label{baseline_eva}

\subsubsection{Knowledge Source Selection} We begin by discussing prompting-based and classification-based methods. This includes presenting several baseline models to select different sources of knowledge for comparison, and we adopt \textbf{F1} as automatic evaluation metrics, following previous works \citep{wang2023large}.

\begin{itemize}[leftmargin=*]
    \item \textbf{Prompting-based methods.} We mainly utilize zero-shot and in-context learning prompting (ICL) to directly instruct ChatGPT (gpt3.5-turbo-1106) and GPT4o (gpt-4o) to output the required knowledge sources (a.k.a, acting tokens). We use same demonstration for fair comparison.
    
    \item \textbf{BERT} \citep{devlin-etal-2019-bert} We utilize BERT as the backbone to train a classifier, which determines the required sources. We formulate this task as a multi-label classification task to determine whether each source is required or not required. If none of them is required, then the used source is \texttt{NULL}. We choose different thresholds for different datasets according to the performance at validation dataset \footnote{In detail, we set the threshold as 0.3 in kBP and 0.5 in DuLeMon datasets, respectively.}.

    \item \textbf{SAFARI} \citep{wang2023large} is the first work to formulate the source planning task as a sequence generation task, regarding different sources of knowledge as different tokens in vocabulary at LLMs. However, it does not consider relevance scores or iterative refinement. We copy the its best performance for better comparison.
\end{itemize}

\subsubsection{Knowledge Retrieval} We compare the performance of retrieval with different types of retrievers, and we choose \textbf{Recall@1} as our primary evaluation metric. This selection is motivated by the predominant use case scenario observed in the original datasets, where a single evidence from each knowledge source is typically sufficient.

\begin{itemize}[leftmargin=*]
    \item \textbf{BM25} \citep{bm25} is a type of sparse retriever, which takes into account term frequencies and inverse document frequency (TF-IDF) to score the relevance of documents to a query.
    
    \item \textbf{RocketQAv2} \citep{ren-etal-2021-rocketqav2} is a type of dense retriever, which is a unified list-wise training approach for both the retriever and the re-ranker.

    \item \textbf{DPR} \citep{dpr} is a method that leverages dense vector representations for passages in a collection. It uses pre-trained language models to encode passages and queries into dense vectors independently, allowing for a more nuanced understanding of the content.

    \item \textbf{LLMs.} We utilize the same instruction as shown in Figure~\ref{llm_sim_prompt} to prompt \texttt{gpt-3.5-turbo-1106} and \texttt{gpt-4o} to retrieve top-n evidence from corresponding source of knowledge. Furthermore, we also employ in-context learning (ICL) prompting for more comprehensive comparison \footnote{We randomly sampled three cases to cover all knowledge sources from the original dataset. We do not observe significant improvements with other demonstration selection strategies.}. 
    
\end{itemize}

\subsubsection{Response Generation} We mainly compare two methods due to limited work focus on our target problem. Regarding the metrics, we choose \textbf{BLEU}, \textbf{Rouge-L} to evaluate the gap between generated response with the ground truth response in the original datasets. Besides that, we select \textbf{Persona Consistency~(P.C)} and \textbf{Knowledge Consistency~(K.C)} to evaluate the consistency score using our finetuned NLI models \cite{madotto-etal-2019-personalizing} with same definition in \cite{wang2023large}. It is important to note we merge the two sources of persona information (\texttt{User-Per}, \texttt{Bot-Per}) in DuLeMon into one unified P.C score.

\begin{itemize}[leftmargin=*]
    \item \textbf{FoCus} \citep{jang2022call} aims to minimize
    the negative log-likelihood of \textit{language modeling} and sub-tasks: \textit{persona grounding} and \textit{knowledge grounding}. It uses either the sigmoid or softmax functions to select evidence by concatenating evidence and dialogue context, and then generate the final response. We do not report the performance on DuLeMon dataset since most of responses in this dataset do not require external sources.

    \item \textbf{SAFARI} \citep{wang2023large} incorporates all retrieved evidences with corresponding source signals to generate the final responses, without considering the relevance between different evidences and dialogue context explicitly. We report performance of both supervised and unsupervised SAFARI.

\end{itemize}

\subsection{Implementation Details}
\label{implementaion_details}

\textbf{UniMS-RAG:} We mainly choose \texttt{ChatGLM-6B} \cite{du2022glm,zeng2023glm-130b} as the backbone models during training, we set the batch size as 8, train models with 3 epochs and save the checkpoint with the lowest validation loss. For other hyper-parameter settings, we mainly follow the corresponding official code \footnote{\url{https://github.com/THUDM/ChatGLM-6B}}. Due to the computation limit, we conduct training with LoRA \cite{hu2021lora} at one single 3090 GPU, and it costs about 4-6 hours. For the prompting using LLMs, we choose \texttt{gpt-3.5-turbo-1106} and \texttt{gpt-4o}, and set both the temperature and top p as 0.1 to reduce the randomness of LLMs. We choose only the top-ranked result for the main experiment to ensure a consistent setting, and we additionally investigate the effects of retrieving different numbers of results to evaluate whether UniMS-RAG can effectively filter out unrelated evidence. According to different sources of similarity scores, we have three variants: \textbf{UniMS-RAG (ChatGPT)}, \textbf{UniMS-RAG (GPT-4o)} and \textbf{UniMS-RAG (DPR)} where the similarity scores come from ChatGPT, GPT4o and DPR, respectively. At the inference stage, we use self-predicted similarity scores to retrieve the evidence, serving as an indicator. More variants or analyses can be found in \S\ref{infer_feedback}.

\noindent \textbf{Others:} We finetune RocketQAv2 and DPR using samples from corresponding dataset by regarding \textit{(context, used\_persona / document)} as the positive and \textit{(context, unrelated\_persona / document)} as the negative. We set epochs as 5 and max sequence length as 512, and mainly follow these codebases\footnote{\url{https://github.com/PaddlePaddle/RocketQA}, \url{https://github.com/Alibaba-NLP/Multi-CPR/tree/main/retrieval}} for other parameters. For RocketQAv2, we load the weights of pre-trained model \texttt{zh\_dureader\_de\_v2} as introduced in the official homepage, which is trained on the largest Chinese QA dataset, and we use 12-layer \texttt{bert-base-chinese} with 110M parameters as backbone model for DPR. We then finetune an NLI model \cite{madotto-etal-2019-personalizing} by regarding \textit{(ground\_persona / document, response)} as the positive and randomly sampled \textit{(unrelated\_persona / document, response)} as the negative. We also use  \texttt{bert-base-chinese} as the backbone model. We concatenate and encode the ground persona/document $k$ and response $r$ in the form of $\mathrm{[CLS]} k \mathrm{[SEP]} r \mathrm{[SEP]}$, and we train the model to predict whether responses are consistent with corresponding personas or documents. The batch size for fine-tuning is 8. The maximal training epoch is 5, and the max sequence length of the encoder is 512. In the experiments, we use the AdamW optimizer with a learning rate of 2e-5 and an epsilon of 1e-6. We evaluate the NLI model on the KBP test set every 500 iterations during training, and we save the checkpoint with the highest performance on the test set. The fine-tuned NLI model achieves $>95\%$ accuracy for both datasets.

\section{Experimental Results}
\label{main_exp}

In this section, we present the performance of our proposed method UniMS-RAG at different sub-tasks, including the final end-to-end generation task.

\begin{table*}[t]
    \centering
    \caption{The F1 of different decisions in \textbf{Planning} of different methods under supervised settings. 1) DuLeMon: There are 1686 \texttt{NULL}, 505 \texttt{USER}, 219 \texttt{BOT} and 6 \texttt{BOTH} in the ground planning; 2) KBP: There are 181 \texttt{NULL}, 125 \texttt{PERSONA} and 923 \texttt{BOTH} in the ground planning.}
    \begin{adjustbox}{max width=0.8\textwidth}
    \begin{tabular}{lccccccc}
    \toprule
    \multirow{2}{*}{\textbf{Model}} &  \multicolumn{4}{c}{\textbf{DuLeMon}}  & \multicolumn{3}{c}{\textbf{KBP}} \\
    \cmidrule(lr){2-5} \cmidrule(lr){6-8}
    & \texttt{NULL} & \texttt{User-Per} & \texttt{Bot-Per} & \texttt{Both} & \texttt{NULL} & \texttt{Persona} & \texttt{Both} \\
    \midrule
    \multicolumn{8}{c}{\textit{Unsupervised Setting}} \\
    \midrule
    \midrule
    ChatGPT-Zero-shot & 1.29 (14) & 23.18 (953) & 0.0 (0) & 0.69 (1449) & 11.45 (116) & 20.67 (233) & 74.88 (880) \\

   ChatGPT-ICL & 1.97 (44) & 34.11 (2075) & 0.0 (0) & 0.66 (297) & 27.95 (699) & 23.14 (238) & 41.98 (292)   \\

    GPT4o-Zero-shot & 42.39 (1041) & 20.42 (1278) & 0.0 (0) & 0.0 (97) & 31.41 (392) & 25.48 (346) & 58.84 (491) \\
    
    GPT4o-ICL & 57.59 (1478) & 18.03 (893) & 0.0 (0) & 0.0 (45) & 28.16 (728) & 31.17 (414) & 15.24 (87) \\
    
    \midrule
    \multicolumn{8}{c}{\textit{Supervised Setting}} \\
    \midrule
    \midrule
    
    BERT & 77.39 (1658) & 46.17 (279) & 39.25 (479) & 0.0 (0) & 1.1 (2) & \textbf{42.02} (113) & 86.28 (1104) \\
    
    SAFARI & 75.34 (1298) & 24.93 (249) & 39.05 (416) & 1.74 (453) & 47.10 (129) & 31.96 (69) & \textbf{86.59} (1031)  \\
    
    \hdashline
    
    UniMS-RAG & \textbf{85.15} (1992) & \textbf{47.94} (296) & \textbf{41.50} (128) & 0.0 (0) & \textbf{50.35} (252) & 8.82 (11) & 84.81 (966) \\
    \bottomrule
    \end{tabular}
    \end{adjustbox}
    \label{tab:decisions}
\end{table*}

\begin{table}[ht]
\setlength{\belowcaptionskip}{0pt}
    \centering
    \caption{The performance (Recall@1) of \textbf{Retrieval} of different types of retrievers. It is note the UniMS-RAG (w/ DPR) means that the similarity labels come from DPR model during training, and for w/ LLMs, the labels comes from zero-shot prompting. 1) DuLeMon: There are 511 samples require to use \texttt{User-Per} and 225 samples require to use \texttt{Bot-Per}; 2) KBP: There are 125 samples require to use \texttt{PERSONA} and 923 samples require to use \texttt{BOTH}.}
    \begin{adjustbox}{max width=0.48\textwidth}
    \begin{tabular}{lccccc}
    \toprule
    \multirow{2}{*}{\textbf{Model}} &  \multicolumn{2}{c}{\textbf{DuLeMon}}  & \multicolumn{3}{c}{\textbf{KBP}} \\
    \cmidrule(lr){2-3} \cmidrule(lr){4-6}
     & \texttt{User-Per} & \texttt{Bot-Per} & \texttt{PERSONA} & \texttt{Both-PERSONA} & \texttt{Both-DOCUMENTS} \\
    \midrule
    \multicolumn{6}{c}{\textit{Unsupervised Setting}} \\
    \midrule
    \midrule
    BM25 & 15.85 & 24.00 & 36.80 & 48.97 &  15.05 \\
    ChatGPT-Zero-shot & 21.14 & 37.33 & 45.60 & 65.33 & 16.14 \\
    ChatGPT-ICL & 17.42 & 31.11 & 52.00 & 74.11 & 19.28 \\
    
    GPT4o-Zero-shot & 26.81 & \underline{44.00} & 58.60 & 71.83 & 24.16 \\
    GPT4o-ICL & 23.68 & 42.00 & 65.60 & 80.17 & 36.19  \\

    \midrule
    \multicolumn{6}{c}{\textit{Supervised Setting}} \\
    \midrule
    \midrule
    RocketQAv2 & \textbf{42.85} & \textbf{46.22} & \underline{80.00} & \underline{92.31} & \underline{50.49}  \\
    DPR & \underline{30.72} & 40.00 & \textbf{83.20} & \textbf{93.07} & \textbf{51.67} \\
    
    \hdashline
    UniMS-RAG & \\
        \quad w/ DPR & 23.87 & 31.11 & 74.40 & 82.88 & 22.32 \\
        \quad w/ ChatGPT & 21.53 & 33.33 & 42.40 & 67.61 & 14.41 \\
        \quad w/ GPT4o & 24.07 & 42.22 & 46.56 & 72.05 & 16.58 \\
    \bottomrule
    \end{tabular}
    \end{adjustbox}
    \label{tab:retriever}
\end{table}

\subsection{Performance of Planning (RQ1)}

There are different types of planning decisions in the different datasets: DuLeMon (using \texttt{NULL}, \texttt{User-Per}, \texttt{Bot-Per} and \texttt{Both} sources of knowledge) and KBP (using \texttt{NULL}, \texttt{Persona}, and \texttt{Both}). Table~\ref{tab:decisions} demonstrates the F1 of planning using different methods under these two datasets. Generally, for \textit{prompting-based methods}, we can find that two observations: 1) in-context learning (ICL) prompting can not always achieves better performances compared with zero-shot, regardless the dataset and models; and 2) It is not guaranteed that GPT-4 outperforms ChatGPT in all metrics. In detail, GPT4o tends to predict more NULL compared with ChatGPT, especially on DuLeMon dataset. On the other side, for the \textit{supervised methods}, the SAFARI model performed slightly worse than the BERT model on the DuLeMon dataset, but their performance was comparable on the KBP dataset. Furthermore, the UniMS-RAG achieves best performance on 4 out of 7 metrics. Specifically, for KBP, both UniMS-RAG and SAFARI most frequently predict \texttt{Both}, followed by \texttt{NULL}, and then the \texttt{Persona} case, mirroring the frequency of these cases in the original dataset. However, UniMS-RAG predicts significantly more \texttt{NULL} cases than SAFARI. Some of these \texttt{NULL} predictions by UniMS-RAG are actually \texttt{Persona} in the original dataset, leading to a lower number of \texttt{Persona} predictions by UniMS-RAG. In addition, we also found that the original data distribution has a serious impact on the final planning performance. For example, there are less than 0.1\% samples in the training set of DuLeMon requiring \texttt{Both} source of knowledge, resulting in poor performance of all methods at \texttt{Both}. A similar phenomenon is also observed on the KBP dataset. Another reason behind this could be the additional token (i.e., evaluation tokens) introduced during the training stage compared with SAFARI. In general, the planning capability of LLMs still needs to be improved to solve the complex multiple sources planning problem in a dialogue system, particularly when dealing with imbalanced or scarce data sources.

\subsection{Performance of Retrieval (RQ2)}

To investigate the RQ2, we examine different types of retrievers, including unsupervised methods (\textbf{BM25}, \textbf{ChatGPT} and \textbf{GPT4o}) and supervised methods (\textbf{RocketQAv2}, \textbf{DPR}, and our proposed \textbf{UniMS-RAG}), in order to evaluate the retrieval performance, providing the ground-truth planning labels (except \texttt{NULL}). Table~\ref{tab:retriever} presents the Recall@1 (R@1) of the different retrievers. 

\subsubsection{The performance of baselines}
There are several observations can be concluded from the results: 1) GPT4o can achieves better performance than ChatGPT no matter which prompting method is chosen (zero-shot or in-context), and the performance of in-context learning prompting can not always achieves better performance in all datasets. For example, the in-context learning prompting (ICL) performs worse than zero-shot on DuLeMon but better on KBP. We attribute this to the effects of demonstrations and the complexity of different datasets; 2) The overall performance of dense vector (e.g, RocketQAv2 and DPR) mostly is better than ChatGPT and GPT4o, and the performance of ChatGPT is better than sparse retriever (e.g, BM25). The gap of former is bigger than the latter, indicating the large improvement room for current methods using LLMs as retriever, particularly at the context of conversational search; 3) It is observed that DPR performs the best out of these retrievers on KBP datasets while RocketQAv2 performs the best on DuLeMon dataset, revealing the importance of dense retrieval models in this task. We attribute the higher performance of RocketQAv2 on DuLeMon to the similar distribution between DuLeMon dataset and pre-training corpus of RocketQAv2; 4) On KBP datasets, all retrievers performs best at \texttt{Both-PERSONA} and worst at \texttt{Both-DOCUMENTS}, indicating the difficulty to retrieve independent source of knowledge since the semantics between different knowledge from \texttt{Both-DOCUMENTS} are similar to the same underlying persona $p^*$, making them more difficult to distinguish. On the other hand, the performance on DuLeMon is even worse than \texttt{Both-DOCUMENTS} on KBP. We suspect this is due to the semantic gap between the dialogue context and used persona in DuLeMon (as shown in Figure~\ref{intro_examples}).

\subsubsection{The performance of UniMS-RAG} Besides, we also present the performance of UniMS-RAG as a retriever using different similarity signals (DPR, ChatGPT or GPT4o) during the training. We emphasize here that this kind of functionality of UniMS-RAG as retriever is additional bonus due to the introduction of sub-task: relevance score prediction. It does not mean that we must use itself as retriever during inference in practice. Our main focus is to determine whether UniMS-RAG can be used as a retriever. Additionally, if it can be used as a retriever, we aim to evaluate its performance and potential. The results show that UniMS-RAG can be directly used as a retriever, since the performance of all variants of UniMS-RAG are better than sparse retriever (BM25) and some of them even outperforms some prompting-based methods. In detail, UniMS-RAG w/ DPR even achieves better performance than ChatGPT, revealing the great potential of UniMS-RAG as retriever once we use more high-quality signals (the performance of DPR is better than ChatGPT). Furthermore, UniMS-RAG w/ ChatGPT achieves comparable performance with ChatGPT on DuLeMon and KBP datasets. Since the performance of UniMS-RAG w/ GPT4o is much better than UniMS-RAG w/ ChatGPT due to more accurate similarity labels provided by GPT4o, we believe that UniMS-RAG can distill the capabilities of original retriever as much as possible if we can provide more data and more fine-grained signals. Therefore, we can derive the answer to \textbf{RQ2} from this analysis: Large Language Models (LLMs) can serve as retrievers directly, achieving comparable performance compared with original retrievers, showcasing a great potential towards a more powerful unified RAG framework. 

\begin{table}
  \caption{The performance of \textbf{Generation} of different methods. We follow the official code of SAFARI to conduct evaluation on DuLeMon dataset. We bold the highest performance and \underline{underline} the second-best performance.}
  \label{tab:generations}
  \begin{adjustbox}{max width=0.48\textwidth}
  \begin{tabular}{lccccccc}
    \toprule
    \multirow{2}{*}{\textbf{Models}} \hspace{10mm} & \multicolumn{3}{c}{DuLeMon} & \multicolumn{4}{c}{KBP} \\
    \cmidrule(lr){2-4} \cmidrule(lr){5-8}
     & BLEU-1 & Rouge-L & P.C & BLEU-1 & Rouge-L & P.C & K.C \\
    \midrule
    \multicolumn{8}{c}{\textit{Unsupervised Setting}} \\
    \midrule
    \midrule
    SAFARI \cite{wang2023large} & 12.11 & 15.46 & 4.96 & 13.74 & 19.69 & 16.92 & 24.89 \\
    \midrule
    \multicolumn{8}{c}{\textit{Supervised Setting}} \\
    \midrule
    \midrule
    FoCus \cite{jang2022call} & - & - & - & 22.51 & 25.29 & 64.52 & 17.90 \\
    SAFARI \cite{wang2023large} & 7.81 & 8.72 & 51.18 & 23.81 & 26.70 & \textbf{76.99} & \underline{42.39} \\
    \midrule
    \multicolumn{8}{l}{\textit{\textbf{Using itself as retriever}}} \\
    \hdashline
    UniMS-RAG & \\
    
    \quad w/ DPR & 18.43 & 20.32 & 63.18 & \underline{29.65} & \underline{32.48} & \underline{75.51} & 42.07 \\
    \quad w/ ChatGPT & \underline{18.61} & 20.33 & 64.83 & 27.86 & 30.86 & 70.38 & 36.53 \\
    \quad w/ GPT4o & 18.31 & 20.13 & 64.86 & 27.93 & 32.13 & 71.32 & 36.70 \\
    
    \midrule
    \multicolumn{8}{l}{\textit{\textbf{Using independent retriever}}} \\
    
    \hdashline
    
    UniMS-RAG & \\
    
    \quad w/ BM25 & 18.52 & 20.34 & 64.07 & 27.30 & 29.22 & 68.99 & 36.29 \\
    \quad w/ DPR & \textbf{18.95} & \textbf{20.87} & 64.28 & \textbf{29.72} & \textbf{32.85} & 73.56 & \textbf{45.00} \\
    \quad w/ ChatGPT & 18.53 & \underline{20.47} & \underline{65.56} & 29.04 & 31.46 & 72.98 & 39.62 \\
    \quad w/ GPT4o & 18.20 & 20.01 & \textbf{65.95} & 29.27 & 32.07 & 73.13 & 41.42 \\
    
    
  \bottomrule
\end{tabular}
\end{adjustbox}
\end{table}

\subsection{Performance of Generation (RQ3)}
To investigate the performance of UniMS-RAG on the response generation task, we conduct two settings: 1) \textit{using itself as retriever}, which means we use UniMS-RAG to retrieve corresponding evidences from the planned sources of knowledge (i.e, the output of planning step); and 2) \textit{using independent retriever} (i.e, BM25, DPR, ChatGPT, and GPT4o) to retrieve corresponding evidences from the planned sources of knowledge (i.e, the output of planning step). Table~\ref{tab:generations} demonstrates the performance of response generation under both supervised and unsupervised settings.

\subsubsection{The performance of baselines} On the one hand, we can find that supervised methods mostly achieve better performance than unsupervised methods except the BLEU-1 and Rouge-L of unsupervised SAFARI on DuLeMon dataset. We carefully check the outputs of unsupervised SAFARI and find that it tends to plan to use source of knowledge (~70\% using both \texttt{User-Per} and \texttt{Bot-Per}) while most of original test samples do not require any sources of knowledge, resulting in extremely low P.C and higher BLEU-1 and Rouge-L. Furthermore, it is evident that SAFARI outperforms FoCus. We emphasise that FoCus treats knowledge selection as a classification task and optimizes it jointly with response generation tasks, leading to efficiency and scalability issues compared with SAFARI and UniMS-RAG.

\subsubsection{The performance of UniMS-RAG} Referring to Tabel~\ref{tab:decisions} and Table~\ref{tab:retriever}, the performance of the planning step and retrieval step largely affects the results in the final generation step. Specifically, when using itself as retriever, we can find that UniMS-RAG using signals from DPR (w/ DPR) leads to better performance in contrast to ChatGPT (w/ ChatGPT) or GPT4o (w / GPT4o), revealing the effectiveness of better retriever signals. The results of GPT4o is also slightly better than ChatGPT due to more accurate similarity scores no matter using itself as retriever or using independent retriever. The gap between DPR and ChatGPT on DuLeMon is relatively small since most of cases here do not require the involvement of external sources of knowledge. Thus, we decide to load parameters of UniMS-RAG w/ DPR to conduct evaluation when using independent retriever. In detail, the results again validate the effectiveness of better retriever (w/ DPR $>$ w/ GPT4o $>$ w/ ChatGPT $>$ w/ BM25). We find that the performance gap between different retrievers in DuLeMon dataset is pretty small (less than 1\%). We suspect this is highly related to the original data distribution in the test dataset, since most of the samples do not require external persona information. Furthermore, we also find that using independent retriever is mostly better than using itself as retriever except the worst BM25, which is consistent with the findings in the performance of retrieval step. To conclude, it is obvious that our proposed method UniMS-RAG outperforms all other baselines in at least 5 out 7 evaluation metrics no matter using itself as retriever or using independent retriever. Combining the performance of planning, retrieval, and response generation together, we can find that UniMS-RAG is capable of serving as a planner, retriever, and reader in a unified manner, leading to better performance in personalized dialogues.

\subsubsection{Discussion of Unified Manner} Despite the clear performance boost from using independent existing retrievers, we emphasize that the performance gain is not solely due to the better relevance scores provided by these retrievers. It is also significantly attributed to our proposed UniMS-RAG, especially the introduction of relevance score prediction task which makes it can attention on relevant evidences while overlooking noisy ones. This conclusion based on two observations: 1) simply using the existing retriever based on SAFARI cannot achieve state-of-the-art results. We have already selected the best-performing version of SAFARI, and all variants of our proposed UniMS-RAG (i.e., using independent retriever and using itself as retriever) largely outperform it, 2) The performance gap between variants of our proposed method is relatively smaller than the gap between ours with baselines, revealing the great potential and flexibility of UniMS-RAG. In practice, we advocate for the unified training of these three sub-tasks in PerDS, as depicted in UniMS-RAG. During inference, the retriever should be selected on a case-by-case basis.

\section{Analysis and Discussions}

In this section, we choose the best model as shown in our previous experiments to investigate the performance changes under different settings (UniMS-RAG w/ DPR). In detail, we start from the performance of our proposed model with different numbers of retrieved evidence (\S\ref{number_of_evidence}), then we study the effects of introduced self-refinement mechanism during the inference stage by re-evaluating the relationship between generated response with the dialogue context or retrieved evidence (\S\ref{infer_feedback}). We present the ablation study to show the effectiveness and rationale of UniMS-RAG (\S\ref{infer_feedback}), followed by the results of human evaluation.

\subsection{Different Numbers of Retrieved Results}
\label{number_of_evidence}
\begin{figure}[t]
    \begin{subfigure}{0.24\textwidth}
        \includegraphics[width=\textwidth]{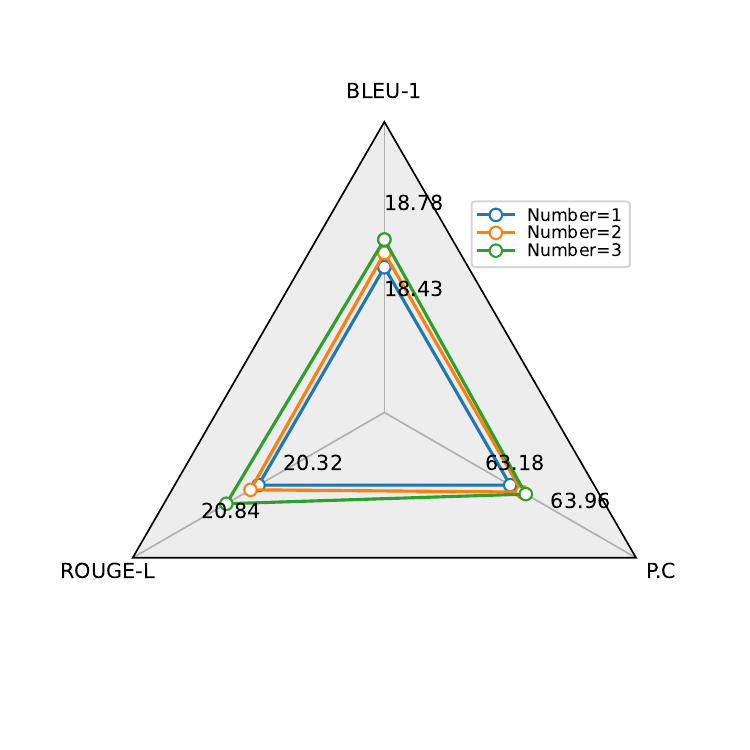}
        \caption{DuLeMon}
        \label{fig:number_of_retrieved_dulemon}
    \end{subfigure}
    \hfill
    \begin{subfigure}{0.22\textwidth}
        \includegraphics[width=\textwidth]{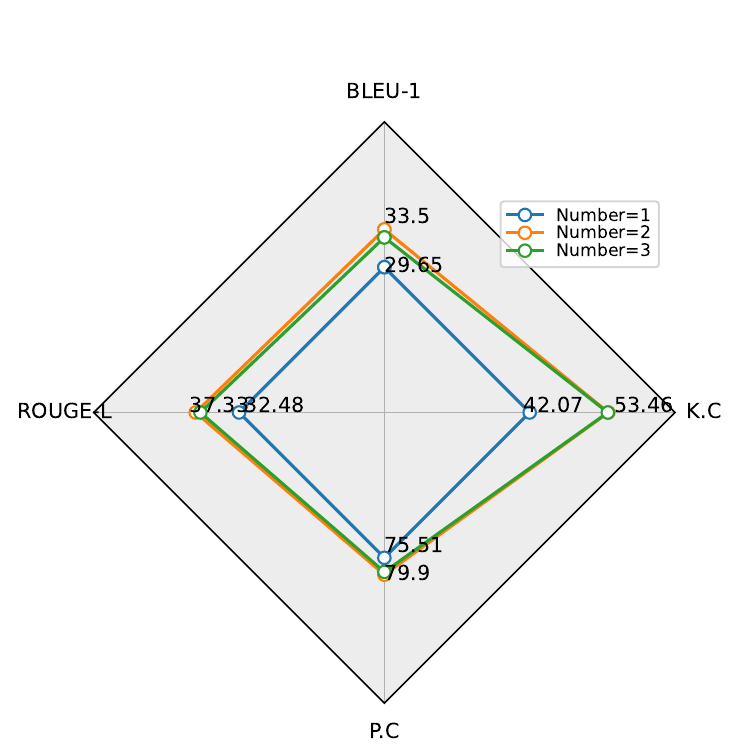}
        \caption{KBP}
        \label{fig:number_of_retrieved_kbp}
    \end{subfigure}

    \caption{The performance of Generation with different number of retrieved evidences on two datasets.}
    \label{fig:number_of_retrieved}
\end{figure}

The number of retrieved results plays a key role in the response generation. Striking a balance between accuracy and recall is essential, as too few results may miss important semantics, while too many can introduce noise. Figure~\ref{fig:number_of_retrieved} shows the performance of UniMS-RAG under different numbers of retrieved results. In DuLeMon dataset, we observe a slight improvement in performance as the number of retrieved results increases. This improvement is likely attributed to infrequent cases requiring evidence in the original test dataset. On the other hand, in KBP dataset, it is obvious that the performance get notably improvement when the number of retrieved evidence increases from 1 to 2, but then experiences a slight decline upon further increase to 3. We hypothesize that this drop is a result of additional noise introduced as the number of retrieved evidence continues to increase. This suggests a delicate balance must be struck to optimize performance, considering the specific characteristics of each dataset.

\subsection{Additional Self-refinement during Inference}
\label{infer_feedback}
To investigate the effects of self-refinement during the inferences, we set the number of retrieved evidence as 3 according to the experimental results in the previous section since it leads to the best performance. There are two different ways to refine the generated response: 1) \textit{Within self-refinement} by providing different number of updated evidences during one-step refinement as shown in Algorithm~\ref{alg:self-refinement}; and 2) \textit{Multi-step self-refinement} by refining the response step by step while keeping the update number fixed. Table~\ref{tab:inference} presents the final results. 

On the one hand, updating the number of evidence within the self-refinement mostly leads to improvement on the consistency scores, e.g., the P.C in DuLeMon, and P.C, K.C in KBP. We suspect this is due to the introduction of novel high-quality and related evidences. Balancing remaining evidence with newly updated evidence introduces a trade-off. If the count of updated evidences surpasses a certain threshold, which unfortunately depends heavily on the number of ground truth evidences employed in each sample, there is a risk of filtering out important information from the remaining evidences. In addition, since we do not refine the response if UniMS-RAG considers it does not require any external sources of knowledge (i.e., \texttt{NULL}), the performance of BLEU-1 and Rouge-L on DuLeMon dataset get slightly drop.

On the other hand, similar patterns emerge when we progressively enhance the response through iterative refinement with an update number of 1, underscoring the efficacy of multi-step self-refinements. However, it becomes evident that the advancement achieved through \textit{multi-step self-refinement }is not uniformly consistent in comparison to \textit{within self-refinement}. This inconsistency is attributed to the interplay between multiple pieces of evidence, given our choice to set the update number as 1. In summary, we posit that integrating these two distinct refinement mechanisms can synergistically augment the quality of generated responses, rendering them more personalized and aligned with existing knowledge.

\begin{table}[!t]
    \centering
    \caption{The performance of \textbf{Generation} with Self-refinement during the inference. Specifically, we first report the performance of different number of updated evidences $\alpha$ (\textit{Within self-refinement}); and then we fix the update number as 1 and iteratively refine the generated response step by step (\textit{Multi-step Self-refinement}).}
    \begin{adjustbox}{max width=0.48\textwidth}
    \begin{tabular}{lccccccc}
    \toprule
   \multirow{2}{*}{\textbf{Number}}  &  \multicolumn{3}{c}{\textbf{DuLeMon}}    &  \multicolumn{4}{c}{\textbf{KBP}}  \\
    \cmidrule(lr){2-4} \cmidrule(lr){5-8}
    & BLEU-1 & Rouge-L & P.C & BLEU-1 & Rouge-L & P.C & K.C \\
    \midrule
    UniMS-RAG & 18.78 & 20.84 & 63.96 & 32.69 & 36.80 & 79.17 & 53.38 \\
    \midrule
    Within Self-refinement \\
    \hdashline
    \quad $\alpha$ = 1 & 18.28 & 20.16 & \textbf{66.50} & 32.60 & 36.76 & 80.23 & 53.45 \\
    \quad $\alpha$ = 2 & 18.25 & 20.34 & 65.48 & 32.99 & 37.34 & 80.47 & 54.43 \\
    \quad $\alpha$ = 3 & \textbf{18.71} & \textbf{20.63} & 65.56 & \textbf{33.33} & \textbf{37.56} & \textbf{81.45} & \textbf{54.84} \\
    \midrule
    Multi-step Self-refinement \\
    \hdashline
    \quad Step = 1 & 18.28 & 20.16 & \textbf{66.50} & 32.60 & 36.76 & 80.23 & 53.45 \\
    \quad Step = 2 & \textbf{18.53} & 20.38 & 65.90 & 33.39 & \textbf{37.48} & 79.09 & 54.52 \\
    \quad Step = 3 & 18.30 & \textbf{20.46} & 65.54 & \textbf{33.52}& 37.35 & \textbf{80.63} & \textbf{54.84} \\
    \bottomrule
    \end{tabular}
    \end{adjustbox}
    \label{tab:inference}
\end{table}


%



\subsection{Ablation Study}
\label{ablation_study}

We investigate the effects of individual steps by providing UniMS-RAG model the ground-truth labels from each step to generate the response, enabling us to analyze and understand the specific effects of each step in a clear and systematic way. Table~\ref{tab:abla} shows the final results.

\begin{table}[!t]
    \centering
    \caption{Ablation study on the impact of different steps and modules in UniMS-RAG.}
    \begin{adjustbox}{max width=0.48\textwidth}
    \begin{tabular}{lccccccc}
    \toprule
    \multirow{2}{*}{\textbf{Model}} &  \multicolumn{3}{c}{\textbf{DuLeMon}}  & \multicolumn{4}{c}{\textbf{KBP}} \\
    \cmidrule(lr){2-4} \cmidrule(lr){5-8} 
    & \textbf{BLEU1} & \textbf{RougeL} & \textbf{P.C} & \textbf{BLEU1} & \textbf{RougeL} & \textbf{P.C} & \textbf{K.C} \\
    \midrule
    UniMS-RAG & 18.43 & 20.32 & 63.18 & 29.65 & 32.48 & 75.51 & 42.07  \\
    \hdashline
    \quad + Ground Planning & 18.27 & 20.43 & 70.92 & 29.82 & 32.73 & 87.88 & 60.21  \\
    \quad + Ground Retrieval & 18.76 & 20.86 & 65.89 & 37.30 & 41.71 & 81.53 & 66.64 \\
    \quad + Ground P \& R & \textbf{19.44}& \textbf{21.89} & \textbf{71.29} & \textbf{37.93} & \textbf{42.67} & \textbf{95.52} & \textbf{87.63} \\
    \midrule
    w/o Attention Mask & 18.32 & 20.20 & 62.70 & 28.69 & 31.44 & 74.53 & 40.93 \\
    \midrule
    w/o Planning \\
    \hdashline
    \quad always use all sources & 17.45 & 19.29 & 1.23 & 29.47 & 32.64 & 73.80 & 36.21 \\
    \quad always do not use source & 17.40 & 19.20 & 69.78 & 26.99 & 28.91 & 14.73 & 24.89 \\
    \bottomrule
    \end{tabular}
    \end{adjustbox}
    \label{tab:abla}
\end{table}

\noindent \textbf{Upper bound of UniMS-RAG.} 
First, we note that the inclusion of ground-truth planning labels or retrieval results casts a positive impact on performance. Planning primarily enhances consistency scores, while grounding retrieval results contributes to BLEU1 and Rouge-L scores. The best results are obtained when both signals are combined, serving as the upper bound of our proposed UniMS-RAG.

\noindent \textbf{The effects of evidence attention mask.} 
There are two notable advantages of adding the evidence attention mask: 1) removing the noise of irrelevant evidence to predict the similarity score; and 2) the UniMS-RAG can effectively operate as a retriever, as it directly captures the relationship between individual contexts and corresponding evidence, optimizing its retrieval capabilities. We can clearly find that removing the evidence attention mask leads to performance degradation, especially on KBP dataset since there are more cases that require external sources of knowledge.

\noindent \textbf{Compared with simple planning strategies.}
We additionally conduct experiments by adopting two simple planning strategies: 1) always use all sources, User-Per and Bot-Per in DuLeMon, and Both in KBP; and 2) always do not use sources, to investigate the effectiveness of the introduced planning step. The results show that the performance is decreased by 63.18\%-1.23\% on DuLeMon and 42.07\%-36.21\% on KBP respectively when using all sources. This highlights that indiscriminate use of all sources not only significantly hampers performance when the majority of samples do not necessitate external references (DuLeMon), but also results in a decline even when a substantial portion of samples requires the utilization of sources (KBP). Conversely, a parallel trend is observed when abstaining from using any external sources. When the majority of samples do not require external references, refraining from their usage leads to notable improvement. However, this approach adversely impacts performance when external sources are essential.

\subsection{Human Evaluation}
\label{human_eva}

Human evaluation is conducted to evaluate the quality of generated response in terms of three metrics: coherence score (\textbf{Coh.}), persona consistency score (\textbf{Per.Cs} for both DuLeMon and KBP), and knowledge consistency score (\textbf{Know.Cs} only for KBP). We randomly sample 100 responses from each dataset with grounding information for each model (We choose UniMS-RAG w/ DPR or w/ ChatGPT and using itself as retriever) and ask three well-educated annotators\footnote{The human inspection and annotation were conducted by a reputable data annotation company, and the annotators are compensated fairly based on the market price without revealing any personal information} to indicate its coherence score (1-5) and whether the response is consistent with the given persona (1/0), and knowledge (1/0). Table~\ref{tab:human_eval} shows the final result. When the number of retrieved evidence is one, we observe there is a significant improvement on DuLeMon and a slight improvement on KBP datasets when using UniMS-RAG w/ DPR. Moreover, when the number of retrieved evidence increases, it is obvious that the performance is largely improved. This observation reveals the effectiveness and robustness of UniMS-RAG to denoise unrelated information, and capture relevant evidences, thanks to the introduction of task and unique training strategies. In addition, UniMS-RAG w/ ChatGPT performs slightly worse than UniMS-RAG w/ DPR due to worse similarity signals. We also observe that human is more likely to find persona-inconsistent cases \citep{wang2023large}. There are some responses that have intra-sentence or intra-context contradictions \citep{zheng-etal-2022-cdconv}, for example, the dialogue responds: ``Yes, our license plate starts with `Yun A'" to the last user turn: ``Does your license plate start with `Yun B'?".

\begin{table}[!t]
    \centering
    \caption{The results of human evaluation. The inter-agreement is about 82\%.}
    \begin{adjustbox}{max width=0.48\textwidth}
    \begin{tabular}{lccccc}
    \toprule
    \multirow{2}{*}{\textbf{Model}} &  \multicolumn{2}{c}{\textbf{DuLeMon}}  & \multicolumn{3}{c}{\textbf{KBP}} \\
    \cmidrule(lr){2-3} \cmidrule(lr){4-6}
    & \textbf{Coh.} & \textbf{Per.Cs (\%)} & \textbf{Coh.} &  \textbf{Per.Cs (\%)}  & \textbf{Know.Cs (\%)} \\
    \midrule
    FoCus & - & - & 3.53 & 54.2 & 33.8   \\
    SAFARI & 2.84 & 14.0 & 4.06 & 68.0 & 59.1 \\
    \midrule
    UniMS-RAG  & & \multicolumn{4}{r}{Number of retrieved evidence = 1}
     \\
     \hdashline
    \quad w/ DPR & 3.22 & 25.0 & 4.18 & 68.1 & 60.5  \\
    \quad w/ ChatGPT & 3.35 & 23.0 & 4.10 & 65.2 & 54.3 \\
    \midrule
    \multicolumn{6}{r}{Number of retrieved evidence = 3}
     \\
     \hdashline
    \quad w/ DPR & 3.38 & 62.0 & 4.20 & 84.8 & 66.7  \\
    \quad w/ ChatGPT & 3.46 & 61.3 & 4.15 & 83.2 & 61.4 \\
    \bottomrule
    \end{tabular}
    \end{adjustbox}
    \label{tab:human_eval}
\end{table}

\section{Conclusion}
In this paper, we focus on personalized knowledge-grounded dialogue tasks in a multi-source setting and decompose the problem into three sub-tasks: \textit{knowledge source selection}, \textit{knowledge retrieval} and \textit{response generation}. We discern a notable gap in none of the existing literature concerning the multiple sources planning and auto-regressive retriever with LLMs themselves as backbone. To fill this gap, we propose a Unified Multi-Source Retrieval-Augmented Dialogue System (UniMS-RAG), aiming to build a unified personalized dialogue system with LLM serving as planner, retriever and reader simultaneously. Experimental results on two popular personalized datasets show that the UniMS-RAG framework can generate more personalized and factual responses and establish a better performance with self-refinement during inference, significantly outperforming strong baseline models under both automatic and human evaluations.


%



\section*{Acknowledgment}

This paper was partially supported by grants from the RGC General Research Funding Scheme (GRF) 14222922 (CUHK 2151185). Work done when Hongru Wang is visiting EdinburghNLP.

\ifCLASSOPTIONcaptionsoff
  \newpage
\fi



%

\bibliographystyle{IEEEtran}
\bibliography{shorter}




%








\end{document}